# Copula-based Kernel Dependency Measures

**Barnabás Póczos**  BAPOCZOS@CS.CMU.EDU
Carnegie Mellon University, School of Computer Science, 5000 Forbes ave, Pittsburgh, PA 15213 USA

**Zoubin Ghahramani**  ZOUBIN@ENG.CAM.AC.UK
University of Cambridge, Department of Engineering, Trumpington Street, Cambridge CB2 1PZ, UK

**Jeff Schneider**  SCHNEIDE@CS.CMU.EDU
Carnegie Mellon University, School of Computer Science, 5000 Forbes ave, Pittsburgh, PA 15213 USA

## Abstract

The paper presents a new copula based method for measuring dependence between random variables. Our approach extends the Maximum Mean Discrepancy to the copula of the joint distribution. We prove that this approach has several advantageous properties. Similarly to Shannon mutual information, the proposed dependence measure is invariant to any strictly increasing transformation of the marginal variables. This is important in many applications, for example in feature selection. The estimator is consistent, robust to outliers, and uses rank statistics only. We derive upper bounds on the convergence rate and propose independence tests too. We illustrate the theoretical contributions through a series of experiments in feature selection and low-dimensional embedding of distributions.

## 1. Introduction

Measuring dependence between random variables is an important problem in statistics, information theory, and machine learning with a wide range of applications in science and engineering. The most well-known dependence measure is the Shannon mutual information, which has found numerous applications recently. Although this is the most popular dependence measure, it is only one of the many other existing ones. In particular, it is a special case of the Rényi-$\alpha$ (Rényi, 1961) and Tsallis-$\alpha$ mutual information (Tsallis, 1988).

Other interesting dependence measures include the maximal correlation coefficient (Rényi, 1959), kernel mutual information (Gretton et al., 2003), the generalized variance and kernel canonical correlation analysis (Bach, 2002), the Hilbert-Schmidt independence criterion (Gretton et al., 2005), the Schweizer-Wolff measure (Schweizer & Wolff, 1981), and the distance based correlation (Székely et al., 2007).

There is a tremendous list of dependence applications. They have been used, for example, in causality detection, feature selection, active learning, structure learning, boosting, image registration, independent component and subspace analysis. For more applications and references, please see the supplementary material.

One reason why so many dependence measures have been defined in the literature is that the problem is challenging and researchers and practitioners are not satisfied with the available measures and estimators (Fernandes & Gloor, 2010). As Schweizer & Wolff (1981) formalized in their dependence axioms, a good dependence measure $I$ has to have several properties. The most important ones are as follows. (i) Dependence $I(\mathbf{X})$ is defined for $\mathbf{X} = (X_1,\ldots,X_d) \in \mathbb{R}^d$ $d$-dimensional random variables. (ii) $I(X_1,\ldots,X_d)$ is invariant to permutation. (iii) $0 \leq I(\mathbf{X})$, and $I(\mathbf{X}) = 0$ iff $(X_1,\ldots,X_d)$ are independent variables. (iv) $I(X_1,\ldots,X_d)$ is invariant to strictly increasing transformation of $X_i$ variables. For more discussion on these axioms, see the Appendix. Among the above mentioned dependence measures, only the Rényi, Tsallis information, and the Schweizer-Wolff measure is invariant to strictly increasing transformations.

In addition to these constraints on the dependence measure, we also want an efficient estimator that is consistent, robust to outliers, has fast convergence rate, and can be used in high-dimensions too. De-pendence estimation is very challenging, especially in nonparametric situations when we cannot assume that the observations have an underlying density function belonging to some parametric family. Many of the above mentioned dependence measures can be defined as some functionals of the density, thus an obvious way for their estimation would be to estimate the densities first. The density function, however, is a nuisance parameter in our case, and its estimation—especially in higher dimensions—is known to be very difficult.

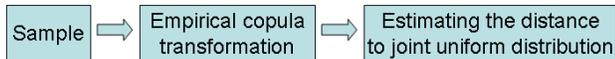

*Figure 1.* Illustration of the proposed dependence measure. Using empirical copula transformation, first we transform the data to have uniform marginals, then measure the distance to the joint uniform distribution with reproducing kernel based divergence estimators.

Due to these difficulties, all the existing dependence estimators have their own shortcomings. For example, the bound on the convergence rate of the Rényi and Tsallis information estimator (Pál et al., 2010) suffers from the curse of dimensionality. The available reproducing kernel based dependence measures are not invariant to strictly increasing transformation of the $X_i$ marginal random variables. The estimator of Székely et al. (2007) is not robust; one single large enough outlier can arbitrarily ruin the estimator.

*The main contributions of the paper* are as follows. (i) We introduce a new dependency measure $I$ that satisfies the above listed axioms. (ii) We prove that $I$ can be efficiently estimated, and the calculation of the estimator is simple. The estimator is consistent, robust to outliers, and uses rank statistics only. (iii) We also provide an upper bound on the rate of convergence and derive a test of independence. This bound shows that the estimator can be efficiently used in large dimensions too.

Our main idea is to combine empirical copula transformations with reproducing kernel based divergence estimators. We will show that the empirical copula transformation only slightly affects the convergence rate, but the resulting dependence estimator possesses all the above mentioned required properties. The proposed method is illustrated in Figure 1.

One might wonder why it is important for a dependence measure to be invariant to strictly increasing transformations of the marginal variables. One reason for this is that in many scenarios we need to compare the estimated dependencies. This is the case for example in feature selection and low-dimensional em-bedding of random variables. In these problems we can think of dependence as a "distance" between random variables in the sense that when the dependence is large, then the random variables are "close" to each other, and when the dependence is small, then the variables are far. However, if certain variables are measured on different scales, then this distance can be much different from the distance using other scales. As a result, it might happen that different features would be selected by the feature selection algorithm if we measured a quantity e.g. in grams, kilograms, pounds, or if we used log-scale. This is an odd situation that can be avoided with dependence measures that are invariant to strictly increasing transformations of the marginal variables. As an application, we will show how the proposed dependence measure can be used for feature selection and low-dimensional embedding of distributions.

The proofs can be found in the supplementary material. There we also discuss the robustness properties of the estimators and show how to use them in independence tests.

**Notation**: In the rest of the paper $X \sim P$ will denote that the random variable $X$ has distribution $P$. $\mathbb{E}(X)$ and $\sigma(X)$ stand for the expectation and standard deviation of $X$, respectively. For a random variable $X \in \mathbb{R}$, $\Xi[X]$ denotes the standardized variable, that is, $\Xi[X] \doteq (X - \mathbb{E}[X])/\sigma(X)$, which has zero mean and unit variance. $U[a, b]$ stands for the uniform distribution in the interval $[a, b]$. $X_{1:m}$ is shorthand notation for the set of random variables $\{X_1, \ldots, X_m\}$. The cardinality of a set $S$ is denoted by $|S|$.

## 2. Maximum Mean Discrepancy

In this section we review some important properties of the Maximum Mean Discrepancy (MMD), which is a quantity used to measure the distance between distributions (Borgwardt et al., 2006; Fortet & Mourier, 1953). An appealing property of this quantity is that it can be efficiently estimated from independent and identically distributed (i.i.d.) samples.

**Definition 1.** *Let $\mathcal{F}$ be a class of functions, $P$, $Q$ be probability distributions. The MMD between $P$ and $Q$ on the function class $\mathcal{F}$ is defined as follows,*

$$\mathcal{M}[\mathcal{F}, P, Q] \doteq \sup_{f \in \mathcal{F}} (\mathbb{E}_{\mathbf{X} \sim P}[f(\mathbf{X})] - \mathbb{E}_{\mathbf{Y} \sim Q}[f(\mathbf{Y})]).$$

Let $\mathcal{H} = \{f : \mathcal{X} \to \mathbb{R}\}$ be a reproducing kernel Hilbert Space (RKHS) with feature map $\phi(x) \in \mathcal{H}$ ($x \in \mathcal{X}$), and kernel $k(x, y) = \langle \phi(x), \phi(y) \rangle_{\mathcal{H}}$. It is well known that $\phi(x) = k(\cdot, x)$, and $f(x) = \langle f, \phi(x) \rangle_{\mathcal{H}}$, which is called the reproducing property of the RKHS. Later we will also need the definition of universal kernels.

**Definition 2** (Universal kernel). *A kernel $k : \mathcal{X} \times \mathcal{X} \to \mathbb{R}$ is universal whenever the associated RKHS $\mathcal{H}$ is dense in $C(\mathcal{X})$, the space of bounded continuous functions over $\mathcal{X}$, with respect to the $L_\infty$ norm.*

Steinwart (2001) has shown that the Gaussian and Laplace kernels are universal. Let $\mu[P] \doteq \mathbb{E}_{\mathbf{X} \sim P}[\phi(\mathbf{X})] = \mathbb{E}_{\mathbf{X} \sim P}[k(\cdot, \mathbf{X})]$. A sufficient condition for this quantity to exist is $\mathbb{E}_{\mathbf{X} \sim P, \mathbf{X}' \sim P} k(\mathbf{X}, \mathbf{X}') < \infty$, where $\mathbf{X}$ and $\mathbf{X}'$ are independent variables having distribution $P$.

For general $\mathcal{F}$ function sets, $\mathcal{M}[\mathcal{F}, P, Q]$ can be difficult to calculate and is not even symmetric in $P$ and $Q$. Nonetheless, when $\mathcal{F}$ is a unit ball of RKHS $\mathcal{H}$, then for all $f \in \mathcal{F}$ we also have that $-f \in \mathcal{F}$, which implies that $\mathcal{M}[\mathcal{F}, P, Q] = \mathcal{M}[\mathcal{F}, Q, P]$. Furthermore, in this case $\mathcal{M}^2[\mathcal{F}, P, Q]$ has a simple form that makes efficient estimations possible. This is stated formally in the following lemma (Borgwardt et al., 2006).

**Lemma 3.** *When $\mathcal{F}$ is a unit ball of RKHS $\mathcal{H}$ and $\mu[P] < \infty$, $\mu[Q] < \infty$, then*

$$\mathcal{M}^2[\mathcal{F}, P, Q] = \|\mu[P] - \mu[Q]\|_{\mathcal{H}}^2 = \mathbb{E}_{\mathbf{X}, \mathbf{X}' \sim P}[k(\mathbf{X}, \mathbf{X}')] \\ - 2\mathbb{E}_{\mathbf{X} \sim P, \mathbf{Y} \sim Q}[k(\mathbf{X}, \mathbf{Y})] + \mathbb{E}_{\mathbf{Y}, \mathbf{Y}' \sim Q}[k(\mathbf{Y}, \mathbf{Y}')],$$

*where $\mathbf{X}$ and $\mathbf{X}'$ have distribution $P$, $\mathbf{Y}$ and $\mathbf{Y}'$ have distribution $Q$, and these random variables are all independent from each other.*

In the remainder of the paper we will always assume that $\mathcal{F}$ is a unit ball of RKHS $\mathcal{H}$. Let $\mathbf{X}_{1:m} = (\mathbf{X}_1, \ldots, \mathbf{X}_m)$ be an independent and identically distributed (i.i.d.) sample drawn from distribution $P$, and similarly let $\mathbf{Y}_{1:n} = (\mathbf{Y}_1, \ldots, \mathbf{Y}_n)$ be an i.i.d. sample with distribution $Q$.

A biased (but asymptotically unbiased) estimator for $\mathcal{M}[\mathcal{F}, P, Q]$ can be easily given using the law of large numbers:

$$\mathcal{M}_b[\mathcal{F}, \mathbf{X}_{1:m}, \mathbf{Y}_{1:n}] \doteq \left[ \frac{1}{m^2} \sum_{i,j=1}^{m} k(\mathbf{X}_i, \mathbf{X}_j) \right. \quad (1)$$
$$\left. + \frac{1}{n^2} \sum_{i,j=1}^{n} k(\mathbf{Y}_i, \mathbf{Y}_j) - \frac{2}{mn} \sum_{i,j=1}^{m,n} k(\mathbf{X}_i, \mathbf{Y}_j) \right]^{1/2}.$$

An unbiased estimator for $\mathcal{M}^2[\mathcal{F}, P, Q]$ (when $m = n$) has also been derived in Borgwardt et al. (2006):

$$\mathcal{M}_u^2[\mathcal{F}, \mathbf{X}_{1:m}, \mathbf{Y}_{1:m}] = \frac{1}{m(m-1)} \sum_{i,j} h(\mathbf{\Lambda}_i, \mathbf{\Lambda}_j), \quad (2)$$

which is a one sample $U$-statistic with $h(\mathbf{\Lambda}_i, \mathbf{\Lambda}_j) \doteq k(\mathbf{X}_i, \mathbf{X}_j) + k(\mathbf{Y}_i, \mathbf{Y}_j) - k(\mathbf{X}_i, \mathbf{Y}_j) - k(\mathbf{X}_j, \mathbf{Y}_i)$, where $\mathbf{\Lambda}_i \doteq (\mathbf{X}_i, \mathbf{Y}_i)$, and $\mathbf{\Lambda}_{1:m} = (\mathbf{\Lambda}_1, \ldots, \mathbf{\Lambda}_m)$ are i.i.d. random variables. From the r.h.s. of Lemma 3, one can see that $\mathbb{E}[h(\mathbf{\Lambda}_i, \mathbf{\Lambda}_j)] = \mathcal{M}^2[\mathcal{F}, P, Q]$, which proves the unbiasedness of the estimator $\mathcal{M}_u^2[\mathcal{F}, \mathbf{X}_{1:m}, \mathbf{Y}_{1:m}]$.

## 3. The Copula of Distributions

Below we review a few important properties of the copula of multivariate distributions that we will use in our work (Nelsen, 1998).

The copula plays an important role when we study the dependence among random variables. The marginal variables $X^1, \ldots, X^d$ are independent from each other, if and only if the copula distribution is the multivariate uniform distribution. In turn, we can measure the dependence of the $X^1, \ldots, X^d$ random variables by measuring how far the copula distribution is from the uniform distribution. The copula contains all the information that we need to measure dependence, and it is invariant to any nonlinear strictly increasing transformations of the marginal variables.

The copula can be defined by the Sklar's theorem (Sklar, 1959) as follows. Let $\mathbf{X} = (X^1, \ldots, X^d) \in \mathbb{R}^d$ be a random variable. Denote the marginal cdf's of $X^j$ by $F_j : \mathbb{R} \to [0, 1]$. Sklar's theorem states that a multivariate cumulative distribution function $H(x_1, \ldots, x_d) = \Pr(X^1 \leq x_1, \ldots, X^d \leq x_d)$ can be written as $H(x_1, \ldots, x_d) = C(F_1(x_1), \ldots, F_d(x_d))$, where $C$ is a unique distribution function on the range of the $F_i$ cdf functions. This distribution function is called the copula of the joint distribution $H$. The distribution of the copula $C$ is the same as the joint distribution of $\mathbf{Z} = (Z^1, \ldots, Z^d) \doteq \mathbf{F}(\mathbf{X}) = (F_1(X^1), \ldots F_d(X^d)) \in \mathbb{R}^d$ random variables. When the $F_i$ cumulative distribution functions are invertible, then $\mathbf{F}(\mathbf{X})$ have uniformly distributed marginal distributions on $[0, 1]$, and the copula distribution can be calculated as $C(y_1, \ldots, y_d) = H\left(F_1^{-1}(y_1), \ldots, F_d^{-1}(y_d)\right)$, where $0 \leq y_i \leq 1$. The relation of the joint distribution $H$, marginal distributions $F_i$, and copula $C$ is illustrated in Figure 2.

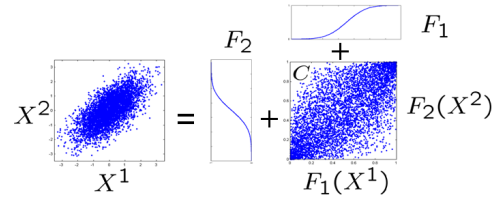

*Figure 2.* Illustration of the copula. On the left: random samples from a 2-dimensional distribution $H$. On the right: the copula transformed sample points. They are distributed according to the copula $C$. Every distribution function $H$ can be rewritten with its copula distribution $C$ and marginal distributions $F_1, F_2$ as $H(X^1 \leq x_1, X^2 \leq x_2) = C(F_1(X^1 \leq x_1), F_2(X^2 \leq x_2))$. The copula $C$ captures all the dependence between $X^1$ and $X^2$. The marginal variables, $X^1$ and $X^2$, are independent iff the copula distribution $C$ is the uniform distrbution.

## 4. Dependence Estimation

Let $\mathbf{U} = (U^1, \ldots, U^d) \in [0,1]^d$ be a random variable with uniform distribution on the $d$-dimensional unit cube, $\mathbf{U} \sim U[0,1]^d$. We define the dependence among continuous random variables $X^1, \ldots, X^d$ as the *MMD distance between the joint copula and the $d$-dimensional uniform distribution*:

$$I(X^1, \ldots, X^d) \doteq \mathcal{M}(\mathcal{F}, P_\mathbf{Z}, P_\mathbf{U}).$$

**Definition 4.** *Let $x_1, x_2 \in \mathbb{R}$. A function $g: \mathbb{R} \to \mathbb{R}$ is strictly increasing, if $g(x_1) < g(x_2)$ for all $x_1 < x_2$.*

It is easy to see that $I(X^1, \ldots, X^d) \geq 0$, and $I(X^1, \ldots, X^d) = I(g_1(X^1), \ldots, g_d(X^d))$ for any $g_i$ strictly increasing functions. In other words, $I(X^1, \ldots, X^d)$ is invariant to strictly increasing transformations of the marginal variables.

The following lemma states that $I(X^1, \ldots, X^d)$ is indeed a well-defined dependence measure when kernel $k$ is universal.

**Lemma 5.** *Let the kernel $k$ be universal on $[0,1]^d \times [0,1]^d$. Then $I(X^1, \ldots, X^d) = 0$, if and only if $X^1, \ldots, X^d$ are independent of each other.*

In what follows we will provide a consistent estimator for $I(\mathbf{X}) = I(X^1, \ldots, X^d)$. Let $k: \mathbb{R}^d \times \mathbb{R}^d \to \mathbb{R}$ be a kernel function of RKHS $\mathcal{H}$, and let $\mathbf{Z} \doteq \mathbf{F}(\mathbf{X})$ be a random variable drawn from the copula. Introduce the following terms:

$$\mu[P_\mathbf{Z}] \doteq \mathbb{E}_{\mathbf{Z} \sim P_\mathbf{Z}}[k((Z^1, \ldots, Z^d), \cdot)],$$
$$\mu[P_\mathbf{U}] \doteq \mathbb{E}_{\mathbf{U} \sim P_\mathbf{U}}[k((U^1, \ldots, U^d), \cdot)].$$

Thanks to Lemma 3, it is easy to see that $I^2(\mathbf{X}) = \mathcal{M}^2(\mathcal{F}, P_\mathbf{Z}, P_\mathbf{U}) = \|\mu[P_\mathbf{Z}] - \mu[P_\mathbf{U}]\|_\mathcal{H}^2 = \mathbb{E}_{\mathbf{Z}, \mathbf{Z}' \sim P_\mathbf{Z}}[k(\mathbf{Z}, \mathbf{Z}')] - 2\mathbb{E}_{\mathbf{Z} \sim P_\mathbf{Z}, \mathbf{U} \sim P_\mathbf{U}}[k(\mathbf{Z}, \mathbf{U})] + \mathbb{E}_{\mathbf{U}, \mathbf{U}' \sim P_\mathbf{U}}[k(\mathbf{U}, \mathbf{U}')]$.

Our goal is to estimate $I(\mathbf{X})$ using the $\mathbf{X}_{1:m}$ i.i.d. sample. This expression is the expected value of the kernel $k$ evaluated in random variables drawn from the uniform and the copula distributions. Assume that we already have a $\mathbf{Z}_{1:m}$ i.i.d. sample from the copula distribution. For simple kernel functions, the expectation w.r.t. the uniform distribution has a simple form. For example, when we use the Gaussian kernel, we have the following unbiased estimator for $I^2(\mathbf{X})$:

$$\mathcal{M}_u^2[\mathcal{F}, \mathbf{Z}_{1:m}, P_\mathbf{U}] = \frac{1}{m(m-1)} \sum_{i \neq j} k(\mathbf{Z}_i, \mathbf{Z}_j)$$
$$- \frac{2}{m} \sum_{i=1}^m \prod_{j=1}^d \int_0^1 \exp\left(\frac{-(\mathbf{Z}_i^j - u)^2}{2\sigma^2}\right) du$$
$$+ \left(\int_0^1 \int_0^1 \exp\left(\frac{-(u-u')^2}{2\sigma^2}\right) du\, du'\right)^d,$$

which can be expressed by the erf Gauss error function. For more complicated kernels, however, these integrals can not be calculated analytically, therefore we need to approximate them by sampling. In what follows we will investigate this case.

Let $\mathbf{U}_{1:n} = \mathbf{U}_1, \ldots, \mathbf{U}_n$ be an i.i.d. sample drawn from the $U[0,1]^d$ distribution, and let $\mathbf{X}, \mathbf{X}_1, \ldots, \mathbf{X}_m$ be i.i.d. samples having distribution $P_\mathbf{X}$. The $F_1, \ldots, F_d$ distribution functions are unknown, but we can estimate them efficiently using the empirical distribution functions. For $x, x^j \in \mathbb{R}$ and $1 \leq j \leq d$, let

$$\widehat{F}_j(x) \doteq \widehat{F}_j(x; X_{1:m}^j) \doteq \frac{1}{m}|\{i: 1 \leq i \leq m, x \leq X_i^j\}|$$
$$\widehat{\mathbf{F}}(x^1, \ldots, x^d) \doteq (\widehat{F}_1(x^1), \ldots, \widehat{F}_d(x^d)) \in \mathbb{R}^d.$$

We call the maps $\mathbf{F}$, $\widehat{\mathbf{F}}$ the *copula transformation*, and the *empirical copula transformation*, respectively. The sample $(\widehat{\mathbf{Z}}_1, \ldots, \widehat{\mathbf{Z}}_m) \doteq (\widehat{\mathbf{F}}(\mathbf{X}_1), \ldots, \widehat{\mathbf{F}}(\mathbf{X}_m)) \in \mathbb{R}^d$ is called the empirical copula (Dedecker et al., 2007). Note that the $j$-th coordinate of $\widehat{\mathbf{Z}}_i$ ($1 \leq i \leq m$) equals

$$\widehat{Z}_i^j = \frac{1}{m}\text{rank}(X_i^j, \{X_1^j, X_2^j, \ldots, X_m^j\}),$$

where $\text{rank}(x, A)$ is the number of elements of $A$ less than or equal to $x$. Also, observe that the random variables $\widehat{\mathbf{Z}}_1, \ldots, \widehat{\mathbf{Z}}_m$ are not even independent. Nonetheless, as we will see from Lemma 7, the empirical copula $(\widehat{\mathbf{Z}}_1, \ldots, \widehat{\mathbf{Z}}_m)$ is a good approximation of an i.i.d. sample $(\mathbf{Z}_1, \ldots, \mathbf{Z}_m) \doteq (\mathbf{F}(\mathbf{X}_1), \ldots, \mathbf{F}(\mathbf{X}_m))$ from the copula distribution of $P_\mathbf{X}$. Using (2), we have that

$$\mathcal{M}_u^2[\mathcal{F}, \mathbf{Z}_{1:m}, \mathbf{U}_{1:m}] = \frac{1}{m(m-1)} \sum_{i \neq j} \Big[ k(\mathbf{Z}_i, \mathbf{Z}_j)$$
$$+ k(\mathbf{U}_i, \mathbf{U}_j) - k(\mathbf{Z}_i, \mathbf{U}_j) - k(\mathbf{U}_i, \mathbf{Z}_j) \Big].$$

From (1), we can also see that

$$\mathcal{M}_b[\mathcal{F}, \mathbf{Z}_{1:m}, \mathbf{U}_{1:n}] = \Bigg[ \frac{1}{m^2} \sum_{i,j=1}^m k(\mathbf{Z}_i, \mathbf{Z}_j)$$
$$+ \frac{1}{n^2} \sum_{i,j=1}^n k(\mathbf{U}_i, \mathbf{U}_j) - \frac{2}{mn} \sum_{i,j=1}^{m,n} k(\mathbf{Z}_i, \mathbf{U}_j) \Bigg]^{1/2}.$$

In these expressions $\mathbf{Z}_{1:m}$ is not available to us. We estimate them using the empirical copula, $\widehat{\mathbf{Z}}_j \doteq \widehat{\mathbf{F}}(\mathbf{X}_j)$, $j = 1, \ldots, m$. An estimator for $I^2(\mathbf{X})$ can be given by $\widehat{I}_u^2(\mathbf{X}_{1:m})$, where

$$m(m-1)\widehat{I}_u^2(\mathbf{X}_{1:m}) \doteq m(m-1)\mathcal{M}_u^2[\mathcal{F}, \widehat{\mathbf{Z}}_{1:m}, \mathbf{U}_{1:m}] =$$
$$\sum_{i \neq j} \Big[ k(\widehat{\mathbf{Z}}_i, \widehat{\mathbf{Z}}_j) + k(\mathbf{U}_i, \mathbf{U}_j) - k(\widehat{\mathbf{Z}}_i, \mathbf{U}_j) - k(\mathbf{U}_i, \widehat{\mathbf{Z}}_j) \Big].$$

To calculate this quantity, we only need the ranks of the marginal variables in the sample. Note that $\widehat{I}_u^2(\mathbf{X}_{1:m})$ is not an unbiased estimator of $I(\mathbf{X})$, but we keep the notation $\widehat{I}_u^2$ to denote that it is derived from the estimator $\mathcal{M}_u^2$.

Using the definition of $\mathcal{M}_b$, we can also propose another estimator for $I(\mathbf{X})$:

$$\widehat{I}_b(\mathbf{X}_{1:m}) \doteq \mathcal{M}_b[\mathcal{F}, \widehat{\mathbf{Z}}_{1:m}, \mathbf{U}_{1:n}] = \left[\frac{1}{m^2} \sum_{i,j=1}^{m} k(\widehat{\mathbf{Z}}_i, \widehat{\mathbf{Z}}_j) \right.$$
$$\left. + \frac{1}{n^2} \sum_{i,j=1}^{n} k(\mathbf{U}_i, \mathbf{U}_j) - \frac{2}{mn} \sum_{i,j=1}^{m,n} k(\widehat{\mathbf{Z}}_i, \mathbf{U}_j) \right]^{1/2}.$$

Both estimators are extremely simple to implement requiring only kernel evaluations on the transformed data and the uniform variables. One can also see that the estimators are robust assuming $k$ is bounded in $[0,1]^d \times [0,1]^d$ (but can be unbounded outside of this region, e.g. polynomial kernel). Thanks to the empirical copula transformation, we only need rank statistics ($\widehat{\mathbf{Z}}_{1:m}$) in the estimation, but the actual values of $\mathbf{X}_{1:m}$ sample points are not used. The contribution of one single sample point is diminishing in the estimator as we increase the sample size. Therefore, one arbitrarily large outlier sample point cannot ruin the statistics arbitrarily badly. For more discussion on this, see the Appendix.

In what follows we will analyze the theoretical properties of these estimators. Assume that the kernel function $k(\cdot, \mathbf{z})$ is uniformly Lipschitz continuous on $[0,1]^d$, i.e. there exists $L > 0$ such that for all $\mathbf{z}, \mathbf{z}_1, \mathbf{z}_2 \in [0,1]^d$ we have that $|k(\mathbf{z}_1, \mathbf{z}) - k(\mathbf{z}_2, \mathbf{z})| \leq L \|\mathbf{z}_1 - \mathbf{z}_2\|$. A typical example is the Gaussian kernel, for which it holds that there exists $L > 0$ such that for all $\mathbf{z}, \mathbf{z}_1, \mathbf{z}_2 \in [0,1]^d$

$$\left| \exp(-\frac{\|\mathbf{z}_1 - \mathbf{z}\|^2}{2\sigma^2}) - \exp(-\frac{\|\mathbf{z}_2 - \mathbf{z}\|^2}{2\sigma^2}) \right| \leq L \|\mathbf{z}_1 - \mathbf{z}_2\|.$$

**Lemma 6.** *For all $\mathbf{z}_i \in [0,1]^d$, $1 \leq i \leq 4$, we have that*

$$|k(\mathbf{z}_1, \mathbf{z}_2) - k(\mathbf{z}_3, \mathbf{z}_4)| \leq L\|\mathbf{z}_1 - \mathbf{z}_3\| + L\|\mathbf{z}_2 - \mathbf{z}_4\|.$$

The effect of the empirical copula transformation can be studied by a version of the classical Kiefer-Dvoretzky-Wolfowitz theorem due to Massart; see e.g. Devroye & Lugosi (2001). As a simple implication of this theorem, one can show that $\widehat{\mathbf{F}}$ is a consistent estimator of $\mathbf{F}$, and the convergence is uniform:

**Lemma 7** (Convergence of the empirical copula). *Let $\mathbf{X}_1, \ldots, \mathbf{X}_m$ be an i.i.d. sample from a probability distribution over $\mathbb{R}^d$ with marginal cdf's $F_1, \ldots, F_d$. Let $\mathbf{F}(\mathbf{X})$ be the copula defined above, and let $\widehat{\mathbf{F}}(\mathbf{X}_{1:m})$ be the empirical copula transformation. Then, for any $\epsilon \geq 0$,*

$$\Pr\left[\sup_{\mathbf{x} \in \mathbb{R}^d} \|\mathbf{F}(\mathbf{x}) - \widehat{\mathbf{F}}(\mathbf{x})\|_2 > \epsilon\right] \leq 2d \exp(-\frac{2m\epsilon^2}{d}).$$

Let $0 \leq k(x,y) \leq K$ be a bounded kernel function. The following theorems state the almost sure consistency of the dependence estimators, and provide upper bounds on the rate of convergence.

**Theorem 8** (Almost sure consistency). *Almost surely we have that*

$$|\widehat{I}_u^2(\mathbf{X}_{1:m}) - I^2(\mathbf{X})|$$
$$= \mathcal{O}\left(\max\left\{\sqrt{\frac{dL^2}{m} \log(4dm^2)}, \sqrt{\frac{K^2}{m} \log(4m^2)}\right\}\right).$$

From the below theorem it follows that when $n$ grows fast enough, then $\widehat{I}_b$ is almost surely consistent as well.

**Theorem 9** (Almost sure consistency). *Let $n = g(m)$ for some function $g$ such that $\lim_{m \to \infty} g(m) = \infty$. Almost surely it holds that*

$$|\widehat{I}_b(\mathbf{X}_{1:m}) - I(\mathbf{X})| = \mathcal{O}\bigg(\max\left\{\left(\frac{8dL^2}{m}\log(4dm^2)\right)^{1/4},\right.$$
$$\left.\left(\frac{2K(m+n)}{mn}\log(4m^2)\right)^{1/2}\right\} + \left(\frac{K}{m}\right)^{1/2} + \left(\frac{K}{n}\right)^{1/2}\bigg).$$

As these bounds show, the proposed dependence estimators can be used in high-dimensions as well; they do not suffer from the curse of dimensionality. Based on these estimators, one can derive independence tests too. For details, see the Appendix.

## 5. Feature Selection

The above defined $I(\mathbf{X})$ dependence measure is invariant to strictly increasing transformations of the marginal variables. In this section we discuss the benefits of this property in the feature selection problem.

Let us have $d$ real valued features $\{X^1, \ldots, X^d\}$, and a target value $Y$. Numerous feature selection methods use dependence estimation for selecting the most relevant features to predict the target value $Y$. If we want to select $h$ features, then one obvious approach would be to select those $h$ features that together have the highest dependence with $Y$. This subset selection problem, unfortunately, is very difficult. Therefore, several approximations and heuristics have been proposed. For example, according to the so-called max-relevance criterion (Peng & Ding, 2005), our goal is to select a feature set $S \subseteq \{X^1, \ldots, X^d\}$, which maximizes the average dependence between the features

and the target:
$$\widehat{S} = \arg\max_S \frac{1}{|S|} \sum_{X^i \in S} I(X^i, Y). \quad (3)$$

This approach might select highly redundant features, i.e. the dependence among these features could be large. This redundancy can be measured by the expression $\sum_{X^i, X^j \in S} I(X^i, X^j)/|S|^2$.

When two features highly depend on each other, then probably we do not lose too much if we remove one of them. Therefore, our goal is to maximize relevance while minimizing the redundancy among the features

$$\widehat{S} = \arg\max_S \sum_{X^i \in S} \frac{I(X^i, Y)}{|S|} - \sum_{X^i, X^j \in S} \frac{I(X^i, X^j)}{|S|^2}. \quad (4)$$

All we need is a good estimator for $I(X^i, X^j)$ and $I(X^i, Y)$ dependencies. Equation (3) and (4) objectives are popular tools for feature selection. Here we will not discuss the advantages and disadvantages of them. We, however, would like to point out that when someone uses objectives that involves dependence estimation, then we want these dependencies to be invariant to strictly increasing transformations of the marginal variables.

## 6. Numerical Illustrations

We illustrate the theoretical contributions of this paper through a series of numerical experiments demonstrating properties of the copula-based kernel dependency measure.

The $\mathcal{M}(\mathcal{F}, P_\mathbf{X}, \prod_{i=1}^d P_{X^i})$ measure could also be used directly, without copula transformation, to estimate dependence. In order to use this approach, we need to generate $m$ sample points from the product distributions of the marginals. Let $\tau_i(1:m)$, $(1 \leq i \leq d)$ denote independent random permutations of $\{1, \ldots, m\}$. Then $\Pi[\mathbf{X}_{1:m}] \doteq (X^1_{\tau_1(1:m)}, X^2_{\tau_2(1:m)}, \ldots, X^d_{\tau_d(1:m)})^T$ can be considered as samples from the $\prod_{i=1}^d P_{X^i}$ distribution. In other words, if $\mathbf{X}_{1:m}$ is stored in a $d \times m$ dimensional sample matrix and we independently permute the elements of each row, then the distributions of the rows (the marginal distributions of $\mathbf{X}$) remain the same, but they become independent from each other. For brevity, we will call the $\mathcal{M}(\mathcal{F}, P_\mathbf{X}, \prod_{i=1}^d P_{X^i})$ quantity MMD dependence measure.

### 6.1. Feature Selection

In this experiment we show that $I(\mathbf{X})$ can achieve better performance in feature selection than MMD without copula transformation ($\mathcal{M}(\mathcal{F}, P_\mathbf{X}, \prod_{i=1}^d P_{X^i})$).

We constructed the following random variables: $X^1 \sim U[0,1]$, $X^2 \sim U[0,500]$, $Y = 500\sin(4\pi X^1)$. The task in this experiment was to choose the feature between $X^1$ and $X^2$ that contains the most information about $Y$. This feature is of course $X^1$ since $Y$ is a deterministic function of it, and $X^2$ is independent of $Y$; it does not contain any information about $Y$. 300 sample points from the joint distrbutions of $(X^1, Y)$ and $(X^2, Y)$ are shown in Figure 3(a) and Figure 3(b), respectively. The empirical copula transformed points of $(Y, X^1)$ and $(Y, X^2)$ are displayed in Figure 3(c) and Figure 3(d). When we simply use MMD without copula transformation ($\mathcal{M}(\mathcal{F}, P_{\mathbf{Y}, X^i}, P_Y \times P_{X^i})$), then interestingly we got that the estimated dependence between $Y$ and $X^1$ ($\mathcal{M}_b(\mathcal{F}, (Y, X^1)_{1:m}, \Pi[(Y, X^1)_{1:m}])$, column (A) of Figure 3(e)) was smaller than the estimated dependence between between $Y$ and $X^2$ ($\mathcal{M}_b(\mathcal{F}, (Y, X^2)_{1:m}, \Pi[(Y, X^2)_{1:m}])$, column (B) of Figure 3(e)). As we can see in this problem, the MMD without copula transformation could not select the right feature. However, when we used copula transformation, then the estimated dependence was larger between $Y$ and $X^1$ than between $Y$ and $X^2$. The values of $\widehat{I}^b((Y, X^1)_{1:m})$ and $\widehat{I}^b((Y, X^2)_{1:m})$ are shown in the (C) and (D) columns of Figure 3(e). In this experiment we used Gaussian kernel with $\sigma = 1$.

### 6.2. Feature Standardization

A frequently used feature preprocessing step is to standardize the features, that is, linearly transform them to have zero mean and unit variance ($\Xi[X]$). One might wonder if this simple transformation can solve the problem of Section 6.1. Below we show an example, where that we have only two zero mean unit variance features, and the MMD feature selection method that is not invariant to the strictly increasing transformations of the features selects a feature that is actually independent from the target value.

Let $U \sim U[0,1]$, $X^1 \doteq \Xi[1/U^2]$, $V \sim U[0,1]$ $X^2 \doteq \Xi[V]$ independent random variables, and let $Y \doteq \Xi[\sin(4\pi X^1)]$. The variables are standardized so they have zero mean and standard variation 1. We sampled 4,000 i.i.d. observations from our observed features $X^1$ and $X^2$. The task again was to select the feature that contains the most information about $Y$. The solution to this problem is $X^1$ again. The meanings of the columns in Figure 4 are the same as in Figure 3(e). When we simply use MMD without copula transformation, then the estimated dependence between between $Y$ and $X^1$ was smaller than between $Y$ and $X^2$ ($\mathcal{M}_b(\mathcal{F}, (Y, X^1)_{1:m}, \Pi[(Y, X^1)_{1:m}])$ and $\mathcal{M}_b(\mathcal{F}, (Y, X^2)_{1:m}, \Pi[(Y, X^2)_{1:m}])$ in column (A) and (B), respectively). The MMD without copula

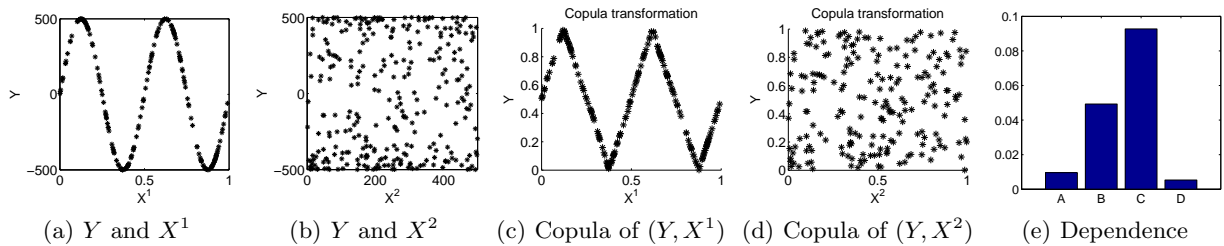

(a) $Y$ and $X^1$  (b) $Y$ and $X^2$  (c) Copula of $(Y, X^1)$  (d) Copula of $(Y, X^2)$  (e) Dependence

Figure 3. (a) Features $Y$ and $X^1$. (b) Features $Y$ and $X^2$. (c) Copula distribution of $(Y, X^1)$. (d) Copula distribution of $(Y, X^2)$. Notations of the bar plot in (e): (A) MMD dependence between $Y$ and $X^1$ ($\mathcal{M}_b(\mathcal{F}, (Y, X^1)_{1:m}, \Pi[(Y, X^1)_{1:m}])$). (B) MMD dependence between $Y$ and $X^2$ ($\mathcal{M}_b(\mathcal{F}, (Y, X^2)_{1:m}, \Pi[(Y, X^2)_{1:m}])$. (C) copula based dependence between $Y$ and $X^1$ ($\widehat{I}^b((Y, X^1)_{1:m})$). (D) copula based dependence between $Y$ and $X^2$ ($\widehat{I}^b((Y, X^2)_{1:m})$).

transformation could not select the right feature. However, when we use copula transformation first, then we can see that the estimated dependence between $Y$ and $X^1$ is larger than between $Y$ and $X^2$, as expected. (C) and (D) show $\widehat{I}^b((Y, X^1)_{1:m})$ and $\widehat{I}^b((Y, X^2)_{1:m})$, respectively.

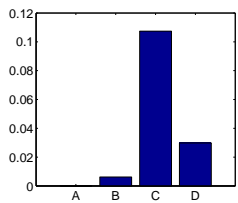

Figure 4. (A-B) Dependence estimation without copula transformation: (A) MMD between $Y$ and $X^1$, $\mathcal{M}_b(\mathcal{F}, (Y, X^1)_{1:m}, \Pi[(Y, X^1)_{1:m}])$, (B) MMD between $Y$ and $X^2$, $\mathcal{M}_b(\mathcal{F}, (Y, X^2)_{1:m}, \Pi[(Y, X^2)_{1:m}])$. (C-D) Dependence estimation with copula transformation: (C) $\widehat{I}^b((Y, X^1)_{1:m})$, (D) $\widehat{I}^b((Y, X^2)_{1:m})$.

### 6.3. Housing Dataset

In the following experiment we study our estimators on the Housing dataset from the UCI repository (Frank & Asuncion, 2010). The dataset contains 506 instances of 14 real valued attributes. The attributes contain various features including per capita crime rate by town, full-value property-tax rate per $10000, average number of rooms per dwelling, percentage of lower status of the population, median value of owner-occupied homes in $1000's, etc. Our goal is to predict some of these attributes and select the most important features for this prediction. Since the dataset contains very different features, it is highly nontrivial how to scale them for feature selection when the applied dependence measure is not invariant to strictly increasing transformations of the marginals. This, however, is not an issue for our proposed dependence measure. In this experiment our goal was to predict the "median value of owner-occupied homes in $1000's" (feature 14) using one single feature. We used $m = n = 300$ instances for training, and the rest of the data for testing. We applied Gaussian kernel ($\sigma^2 = 1/12$) in the estimators. The MMD without copula transformation chose the "average number of rooms per dwelling" (feature 6) as the closest feature. When instead of MMD we used the proposed $\widehat{I}_b$ estimator, it selected the "lower status of the population" (feature 13). To study the prediction errors of the selected features, we trained linear regressors for each feature using them as explanatory variables. The prediction errors on the test data are shown in Figure 5. In this experiment the smallest error was achieved by the feature that $\widehat{I}_b$ selected (feature 13). MMD without the copula transformation selected the feature that gave only the second smallest error (feature 6).

Low-dimensional embedding can help us visualize the pairwise dependence structure of random variables. For each feature $X^i$, $X^j$, we estimated the $d(i, j) = \exp(-I(X^i, X^j))$ quantities. This $d(i, j)$ is large when $X^i$, $X^j$ is independent, and small when the dependence between them is large. We considered them as "distances" (although the triangle inequality does not hold between them), and then applied multi-dimensional scaling to embed them into a 2d space. The Housing dataset was used in this experiment too using the same set-up as in the previous study. To estimate the dependence between the features, we tested again $\widehat{I}_b$ (Figure 6(a)) and MMD without copula transformation (Figure 6(b)). We can observe that the locations of these embedded points are very different. If we applied any strictly increasing transformations to the marginal variables, it would not affect the embedding with copula transformation, but we would get very different results when we use MMD without copula transformation. For more numerical experiments, see the supplementary material.

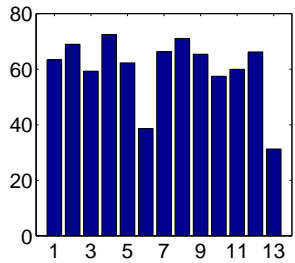

Figure 5. Prediction errors of the features.

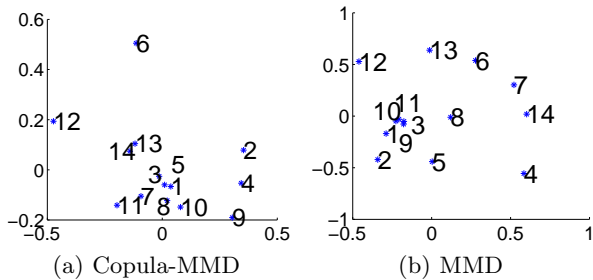

(a) Copula-MMD  (b) MMD

Figure 6. Low-dimensional embedding of the features using dependence as a proximity measure.

## 7. Discussion and Conclusion

We introduced a new RKHS-based dependence measure that operates on the copula of continuous distributions. We have shown that the dependence measure is invariant to strictly increasing transformations of the marginal variables, and this property is important in feature selection and low-dimensional embedding of distributions. We also proposed estimators that are almost surely consistent, robust, use rank statistics only, and do not suffer from the curse of dimensionality. We derived upper bounds on the rates of convergence and illustrated the theory through a series of numerical experiments.